\documentclass{article}
\usepackage{spconf,amsmath,graphicx}
\usepackage{url}
\usepackage{subcaption}
\usepackage{float}% If comment this, figure moves to Page 2
\usepackage{hyperref}
\usepackage{adjustbox}
\title{Convolutional Recurrent Neural Networks for Music Classification}
\twoauthors
  {Keunwoo Choi, Gy\"orgy Fazekas, Mark Sandler\sthanks{This work has been part funded by FAST IMPACt EPSRC Grant EP/L019981/1 and the European Commission H2020 research and innovation grant AudioCommons (688382). Mark Sandler acknowledges the support of the Royal Society as a recipient of a Wolfson Research Merit Award.} }
	{Queen Mary University of London, London, UK\\
	Centre for Digital Music, EECS\\
	E1 4FZ, London, UK\\
	\texttt{keunwoo.choi{@}qmul.ac.uk}
    }
  {Kyunghyun Cho\sthanks{Kyunghyun Cho thanks the support by Facebook, Google (Google Faculty Award 2016) and NVidia (GPU Center of Excellence 2015-2016)}}
	{New York University\\
	Computer Science \& Data Science\\
	New York, NY, USA\\
	\texttt{kyunghyun.cho@nyu.edu}}
\begin{document}
\maketitle
%--------------------------%
% Abstract                 %
%--------------------------%
\begin{abstract}
\vspace{-0.1cm}
We introduce a convolutional recurrent neural network (CRNN) for music tagging. CRNNs take advantage of convolutional neural networks (CNNs) for local feature extraction and recurrent neural networks for temporal summarisation of the extracted features. We compare CRNN with three CNN structures that have been used for music tagging while controlling the number of parameters with respect to their performance and training time per sample. 
%We found CRNNs to outperform baseline networks, indicating the effectiveness of its hybrid structure in music feature extraction and feature summarisation.
Overall, we found that CRNNs show a strong performance with respect to the number of parameter and training time, indicating the effectiveness of its hybrid structure in music feature extraction and feature summarisation.
\end{abstract}
\vspace{-0.1cm}
\vspace{-0.3cm}
%--------------------------%
% Introduction             %
%--------------------------%
\section{Introduction}
\label{sec:intro} 
\vspace{-0.2cm}
Convolutional neural networks (CNNs) have been actively used for various music classification tasks such as music tagging \cite{dieleman2014end, choi2016automatic}, genre classification \cite{sigtia2014improved, chiliguano2016hybrid}, and user-item latent feature prediction for recommendation \cite{van2013deep}. 

%  <--------Figure: Structures ---------> ↓
\begin{figure*}[t!]
  \centering
    \begin{subfigure}[b]{.54\columnwidth}
        \includegraphics[width=1.0\columnwidth]{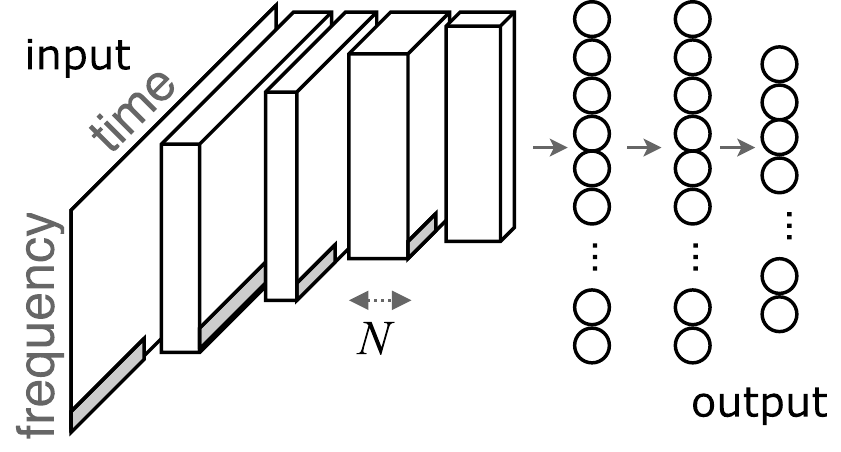}
                                         \vspace{-0.7cm}
        \caption{\texttt{k1c2}}\label{fig:dia_a}
    \end{subfigure} \hspace{0.2cm}
    \begin{subfigure}[b]{.50\columnwidth}
        \includegraphics[width=1.0\columnwidth]{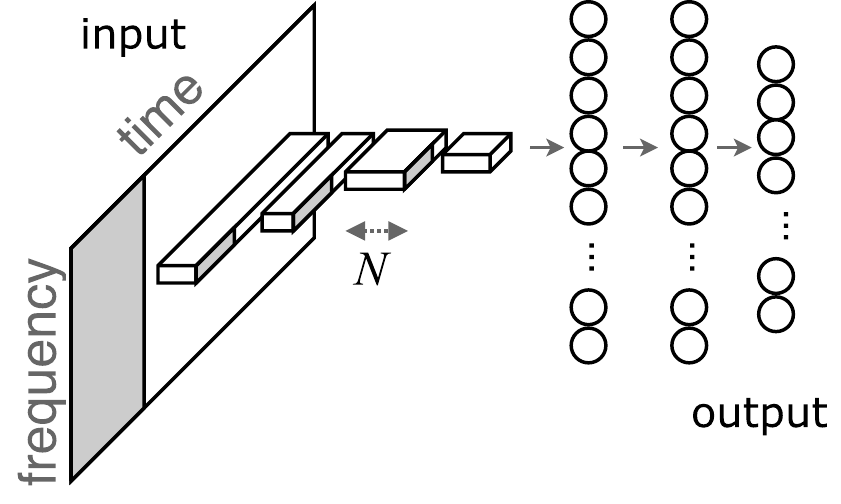}
                                         \vspace{-0.7cm}
        \caption{\texttt{k2c1}}\label{fig:dia_b}
    \end{subfigure} \hspace{0.2cm}
    \begin{subfigure}[b]{.40\columnwidth}
        \includegraphics[width=1.0\columnwidth]{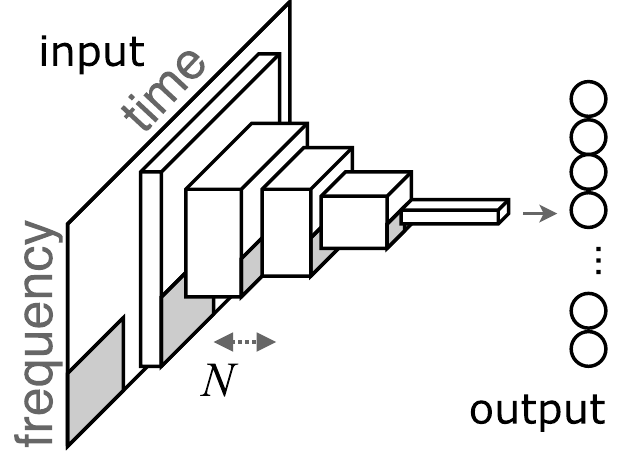}
                                         \vspace{-0.7cm}
        \caption{\texttt{k2c2}}\label{fig:dia_c}
    \end{subfigure} \hspace{0.2cm}
    \begin{subfigure}[b]{.45\columnwidth}
        \includegraphics[width=1.0\columnwidth]{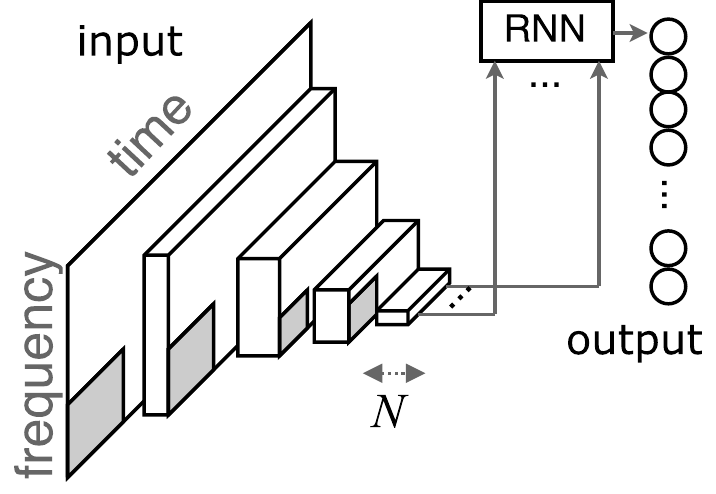}
                                 \vspace{-0.7cm}
        \caption{\texttt{CRNN}}\label{fig:dia_d}
    \end{subfigure}
                         \vspace{-0.2cm}
    \caption{Block diagrams of \texttt{k1c2}, \texttt{k2c1}, \texttt{k2c2}, and \texttt{CRNN}. The grey areas illustrate the convolution kernels. $N$ refers to the number of feature maps of convolutional layers.}
    \label{fig:diagrams}    
\end{figure*}	
%  <--------Figure: Structures ---------> 
CNNs assume features that are in different levels of hierarchy and can be extracted by convolutional kernels. The hierarchical features are learned to achieve a given task during supervised training. For example, learned features from a CNN that is trained for genre classification exhibit low-level features (e.g., onset) to high-level features (e.g., percussive instrument patterns) \cite{choi2016explaining}.

Recently, CNNs have been combined with recurrent neural networks (RNNs) which are often used to model sequential data such as audio signals or word sequences. This hybrid model is called a convolutional recurrent neural network (CRNN). A CRNN can be described as a modified CNN by replacing the last convolutional layers with a RNN. In CRNNs, CNNs and RNNs play the roles of feature extractor and temporal summariser, respectively. Adopting an RNN for aggregating the features enables the networks to take the global structure into account while local features are extracted by the remaining convolutional layers. This structure was first proposed in \cite{tang2015document} for document classification and later applied to image classification \cite{zuo2015convolutional} and music transcription \cite{sigtia2016end}. 

CRNNs fit the music tagging task well. RNNs are more flexible in selecting how to summarise the local features than CNNs which are rather static by using weighted average (convolution) and subsampling. This flexibility can be helpful because some of the tags (e.g., mood tags) may be affected by the global structure while other tags such as instruments can be affected by local and short-segment information. 
% Second, RNNs can effectively deal with variable length of music signal, while CNNs expect a fixed input size.

In this paper, we introduce CRNNs for music tagging and compare them with three existing CNNs. For correct comparisons, we carefully control the hardware, data, and optimisation techniques, while varying two attributes of the structure: \textit{i)} the \textit{number of parameters} and \textit{ii)} \textit{computation time}. 
% In other words, the representation power of the structures is evaluated with respect to memory usage and computation complexity.

%--------------------------%
% Models
%--------------------------%
\vspace{-0.3cm}
\section{Models} \label{sec:mod}
\vspace{-0.2cm}
We compare \texttt{CRNN} with \texttt{k1c2}, \texttt{k2c1}, and \texttt{k2c2}, which are illustrated in Figure \ref{fig:diagrams}. The three convolutional networks are named to specify their kernel shape (e.g., \texttt{k1} for 1D kernels) and convolution dimension (e.g. \texttt{c2} for 2D convolutions). The specifications are shown in Table \ref{table:numbers}. For all networks, the input is assumed to be of size $96$$\times$$1366$ (mel-frequency band$\times$time frame) and single channel. Sigmoid functions are used as activation at output nodes because music tagging is a \textit{multi}-label classification task.

In this paper, all the convolutional and fully-connected layers are equipped with identical optimisation techniques and activation functions -- batch normalization \cite{ioffe2015batch} and ELU activation function \cite{clevert2015fast}. This is for a correct comparison since optimisation techniques greatly improve the performances of networks that are having essentially the same structure. 
%For example, techniques such as batch normalization (BN) \cite{ioffe2015batch} and exponential linear unit (ELU) \cite{clevert2015fast} improve both training speed and accuracy by nudging the distribution of the outputs of intermediate layers to be standardised without changing the structure.
%
Exceptionally, \texttt{CRNN} has weak dropout (0.1) between convolutional layers to prevent overfitting of the RNN layers \cite{srivastava2014dropout}.

% ===== SUBSECTION: k1c1 ===== %
\vspace{-0.4cm}
\subsection{CNN - \texttt{k1c2}}
\vspace{-0.15cm}
\texttt{k1c2} in Figure \ref{fig:dia_a} is motivated by structures for genre classification \cite{li2010automatic}. The network consists of 4 convolutional layers that are followed by 2 fully-connected layers. One-dimensional convolutional layers ($1$$\times$$4$ for all, i.e., convolution along time-axis) and max-pooling layers (($1$$\times$$4$)-($1$$\times$$5$)-($1$$\times$$8$)-($1$$\times$$8$)) alternate. Each element of the last feature map (the output of the 4-th sub-sampling layer) encodes a feature for each band.
%, and resulting in \textit{narrow-and-tall} feature maps
They are flattened and fed into a fully-connected layer, which acts as the classifier.

%This model learns \textit{temporal filters} for each frequency band.
%This model preserves the frequency band information until the end of convolutional layers. In other words, each element of the last feature map (the output of the 4-th sub-sampling layer) encodes a feature for each band. 

% ===== SUBSECTION: k2c1: sander ===== %
\vspace{-0.4cm}
\subsection{CNN - \texttt{k2c1}}
\vspace{-0.15cm}
\texttt{k2c1} in Figure \ref{fig:dia_b} is motivated by structures for music tagging \cite{dieleman2014end} and genre classification \cite{wulfing2012unsupervised}. The network consists of 5 convolutional layers that are followed by 2 fully-connected layers. The first convolutional layer ($96\times4$) learns 2D kernels that are applied to the whole frequency band. After then, one-dimensional convolutional layers ($1$$\times$$4$ for all, i.e., convolution along time-axis) and max-pooling layers (($1$$\times$$4$) or ($1$$\times$$5$)) alternate.
%, and resulting in \textit{narrow-and-tall} feature maps. 
The results are flattened and fed into a fully-connected layer.

%This model relies on the assumption of non-stationarity along the frequency axis of music spectrograms. 
This model compress the information of whole frequency range into one band in the first convolutional layer and this helps reducing the computation complexity vastly. 
%In \cite{dieleman2014end}, 29s-long signal are trimmed into less than 4-sec subsegments, and then the final tag predictions are averaged. The author's more recent work\footnote{\url{http://benanne.github.io/2014/08/05/spotify-cnns.html}} directly takes the whole 30s-long signal as an input. We adopt the approach from the latter in this paper.

% ===== SUBSECTION: CONV2D ===== %
\vspace{-0.3cm}
\subsection{CNN - \texttt{k2c2}}
\vspace{-0.15cm}
CNN structures with 2D convolution have been used in music tagging \cite{choi2016automatic} and vocal/instrumental classification \cite{schluter2016learning}. \texttt{k2c2} consists of five convolutional layers of $3$$\times$$3$ kernels and max-pooling layers (($2$$\times$$4$)-($2$$\times$$4$)-($2$$\times$$4$)-($3$$\times$$5$)-($4$$\times$$4$)) as illustrated in Figure \ref{fig:dia_b}. The network reduces the size of feature maps to $1$$\times$$1$ at the final layer, where each feature covers the whole input rather than each frequency band as in \texttt{k1c1} and \texttt{k2c1}.

This model allows time and frequency invariances in different scale by gradual 2D sub-samplings. Also, using 2D subsampling enables the network to be \textit{fully-convolutional}, which ultimately results in fewer parameters.
% This is because fully-connected layers are often responsible for large portion of the network's parameters.

% ===== SUBSECTION: CRNN ===== %
\vspace{-0.3cm}
\subsection{CRNN}
\vspace{-0.2cm}
\texttt{CRNN} uses a 2-layer RNN with gated recurrent units (GRU) \cite{cho2014properties} to summarise temporal patterns on the top of two-dimensional 4-layer CNNs as shown in Figure \ref{fig:dia_c}. The assumption underlying this model is that the temporal pattern can be aggregated better with RNNs then CNNs, while relying on CNNs on input side for local feature extraction. 

% This structure is first proposed in \cite{tang2015document} for document classification. 
In \texttt{CRNN}, RNNs are used to aggregate the temporal patterns instead of, for instance, averaging the results from shorter segments as in \cite{dieleman2014end} or convolution and sub-sampling as in other CNN's.
In its CNN sub-structure, the sizes of convolutional layers and max-pooling layers are 3$\times$3 and ($2$$\times$$2$)-($3$$\times$$3$)-($4$$\times$$4$)-($4$$\times$$4$). This sub-sampling results in a feature map size of $N$$\times$1$\times$15 (number of feature maps$\times$frequency$\times$time). 
They are then fed into a 2-layer RNN, of which the last hidden state is connected to the output of the network.

% ↓  <--------Table : Architectures ---------> ↓
\begin{table*}[t!]
  \begin{center}
  \setlength\tabcolsep{1.5pt} % default value: 6pt
  \begin{adjustbox}{max width=\textwidth,center}

    \begin{tabular}{ c|c|c|c|c|c || c|c|c|c|c|c || c|c|c|c|c|c || c|c|c|c|c|c } 
    \hline
      \multicolumn{6}{c ||}{\textbf{\texttt{k1c2}}} & \multicolumn{6}{c||}{\textbf{\texttt{k2c1}}} & \multicolumn{6}{c||}{\textbf{\texttt{k2c2}}} & \multicolumn{6}{c}{\textbf{\texttt{CRNN}}}\\ \hline
      \shortstack{No. params \\ ($\times10^6$)}  & 0.1 & 0.25 & 0.5 & 1.0 & 3.0 & & 0.1 & 0.25 & 0.5 & 1.0 & 3.0 & & 0.1 & 0.25 & 0.5 & 1.0 & 3.0 & $$ &0.1 & 0.25 & 0.5 & 1.0 & 3.0\\ \hline \hline
      Layer type  & \multicolumn{5}{c ||}{Layer width} & Type & \multicolumn{5}{c ||}{ Layer width}  &  Type & \multicolumn{5}{c}{Layer width }  \\
      \hline 
            \texttt{conv2d} & 15 & 23 & 33 & 47 & 81& \texttt{conv1d} & 43 & 72 & 106   & 152 & 265 & \texttt{conv2d} & 20 & 33   & 47   & 67   & 118 & \texttt{conv2d} & 30 & 48 & 68   & 96 & 169 \\
            \texttt{conv2d} & 15 & 23 & 33 & 47 & 81& \texttt{conv1d} & 43 & 72 & 106   & 152 & 265 & \texttt{conv2d} & 41 & 66   & 95   & 135 & 236 & \texttt{conv2d} & 60 & 96 & 137 & 195 & 339 \\
            \texttt{conv2d} & 30 & 47 & 66 & 95 & 163& \texttt{conv1d} & 43 & 72 & 106   & 152 & 265 & \texttt{conv2d} & 41 & 66   & 95   & 135 & 236 & \texttt{conv2d} & 60 & 96 & 137 & 195 & 339 \\
            \texttt{conv2d} & 30 & 47 & 66 & 95 & 163&\texttt{conv1d} & 87 & 145 & 212  & 304 & 535 & \texttt{conv2d} & 62 & 100 & 142 & 203 & 355 &  \texttt{conv2d} & 60 & 96 & 137 & 195  & 339 \\
                  \texttt{FC} & 30 & 47 & 66 & 95 & 163 & \texttt{conv1d} & 87 & 145 & 212  & 304  & 535 & \texttt{conv2d} & 83 & 133 & 190 & 271 & 473 & \texttt{rnn}       & 30 & 48  & 68   & 96 & 169\\
                  \texttt{FC} & 30 & 47 & 66 & 95 & 163 &      \texttt{FC}  & 87 & 145 & 212 & 304 & 535 &  $$              &      &        &        &    &     & \texttt{rnn}      & 30 & 48  & 68   & 96 & 169\\
                  $$ & &  & &  &  &                                           \texttt{FC} & 87 & 145 & 212 & 304 & 535 &  $$              &      &        &        &    &     & $$ &  &  &    &  & \\
             \hline              
%      \hline 
%       AUC score & .772 &.781 & .795 & .806 & .829 &$$ & .799 & .814 & .821 & 828 & .857 &$$ & .823 & .838 & .842 & .851 & .855 \\ \hline
%      Time/2.5k samples [s] & 55 & 80 & 111& 146 & 218 & $$ & 46& 64 & 82 & 116 & 180 &$$ & 70 & 103 & 166 & 256 & 402 \\  \hline
    \end{tabular}
    \end{adjustbox}
                     \vspace{-0.15cm}
  \caption{Hyperparameters, results, and time consumptions of all structures. Number of parameters indicates the total number of trainable parameters in the structure. Layer width indicates either the number of feature maps of a convolutional layer or number of hidden units of fully-connected/RNN layers. Max-pooling is applied after every row of convolutional layers. }\label{table:numbers}
  \end{center}
  \vspace{-0.55cm}
\end{table*}
%  <--------Table : Architectures --------->
% ===== SUBSECTION: DEEPER NETWORKS ===== %
\vspace{-0.4cm}
\subsection{Scaling networks}
\vspace{-0.1cm}
The models are scaled by controlling the number of parameters to be 100,000, 250,000, 0.5 million, 1M, 3M with 2$\%$ tolerance. Considering the limitation of current hardware and the dataset size, 3M-parameter networks are presumed to provide an approximate upper bound of the structure complexity. Table \ref{table:numbers} summarises the details of different structures including the \textit{layer width} (the number of feature maps or hidden units).

The widths of layers are based on \cite{dieleman2014end} for \texttt{k1c2} and \texttt{k2c1}, and \cite{choi2016automatic} for \texttt{k2c2}. For \texttt{CRNN}, the widths are determined based on preliminary experiments which showed the relative importance of the numbers of the feature maps of convolutional layers over the number of hidden units in RNNs.

Layer widths are changed to control the number of parameters of a network while the depths and the convolutional kernel shapes are kept constant. Therefore, the hierarchy of learned features is preserved while the numbers of the features in each hierarchical level (i.e., each layer) are changed. This is to maximise the representation capabilities of networks, considering the relative importance of depth over width \cite{eldan2015power}.

%--------------------------%
% Experiments
%--------------------------%
\vspace{-0.3cm}
\section{Experiments} \label{sec:exp}
\vspace{-0.23cm}
We use the Million Song Dataset \cite{bertin2011million} with \textit{last.fm} tags. We train the networks to predict the top-50 tag, which includes genres (e.g., \textit{rock, pop}), moods (e.g., \textit{sad, happy}), instruments (e.g., \textit{female vocalist, guitar}), and eras (\textit{60s -- 00s}). 214,284 (201,680 for training and 12,605 for validation) and 25,940 clips are selected by using the originally provided training/test splitting and filtering out items without any top-50 tags. The occurrences of tags range from 52,944 (\textit{rock}) to 1,257 (\textit{happy}).

We use 30-60s preview clips which are provided after trimming to represent the \textit{highlight} of the song. We trim audio signals to 29 seconds at the centre of preview clips and downsample them from 22.05 kHz to 12 kHz using Librosa \cite{mcfee2015librosa}. Log-amplitude mel-spectrograms are used as input since they have outperformed STFT and MFCCs, and linear-amplitude mel-spectrograms in earlier research \cite{choi2016automatic, dieleman2014end}. The number of mel-bins is 96 and the hop-size is 256 samples, resulting in an input shape of $96$$\times$$1366$.

The model is built with Keras \cite{chollet2015keras} and Theano \cite{team2016theano}. We use ADAM for learning rate control \cite{DBLP:journals/corr/KingmaB14} and binary cross-entropy as a loss function. The reported performance is measured on test set and by AUC-ROC (Area Under Receiver Operating Characteristic Curve) given that tagging is a multi-label classification. Models and split sets are shared online\footnote{\url{https://github.com/keunwoochoi/icassp_2017}}.

We use early-stopping for the all structures -- the training is stopped if there is no improvement of AUC on the validation set while iterating the whole training data once. 

% ↓  ---- Figure - AUC by #params ---- ↓
\begin{figure}[t!]
  \begin{center}
        \includegraphics[width=1.0\columnwidth]{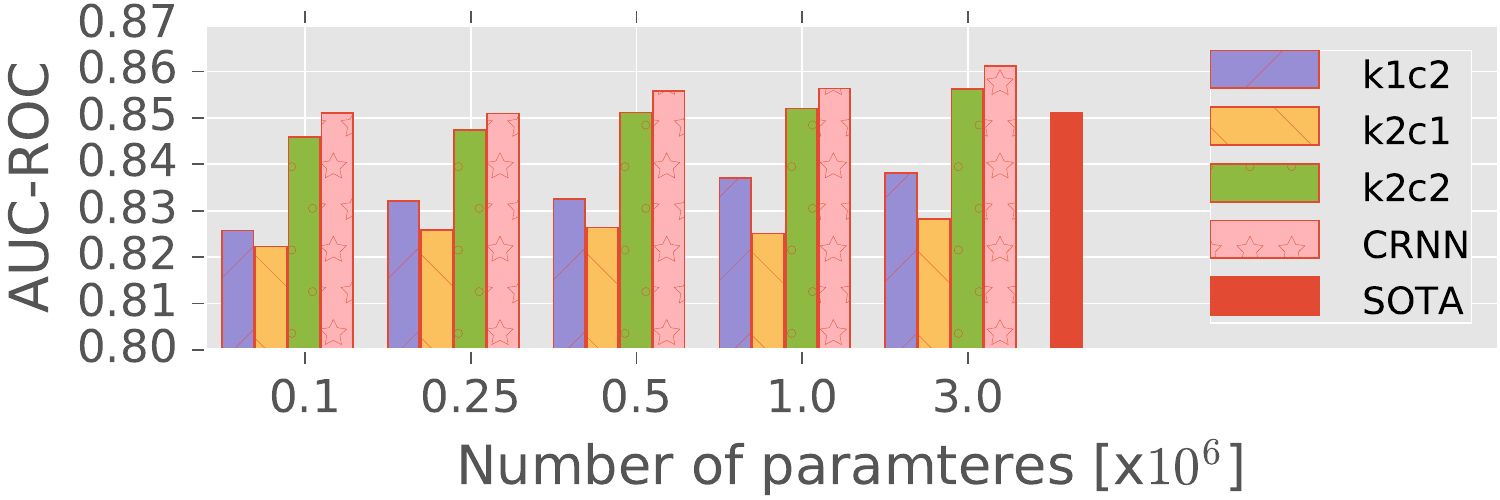}
                     \vspace{-0.73cm}
        \caption{AUCs for the three structures with $\{$0.1, 0.25, 0.5, 1.0, 3.0$\}$$\times$$10^6$ parameters. The AUC of SOTA is .851 \cite{choi2016automatic}.}\label{fig:results}
  \end{center}  
  \vspace{-0.65cm}
\end{figure}
% ↑  ---- FIGURE - AUC by #params ---- ↑

% ↓ ---- FIGURE - AUC per tags ---- ↓
\begin{figure*}[t!]
  \centering
     \includegraphics[width=1.0\textwidth]{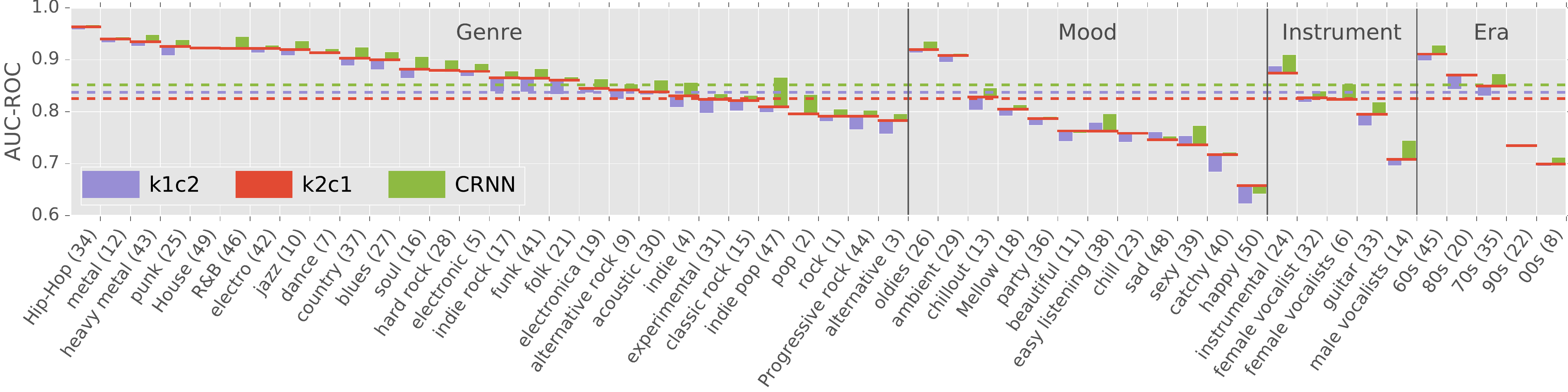}
     \vspace{-0.75cm}     
    \caption{AUCs of 1M-parameter structures. \textit{i}) The average AUCs over all samples are 
    plotted with dashed lines. \textit{ii}) AUC of each tag is plotted using a bar chart and line. For each tag, red line indicates the score of \texttt{k2c1} which is 
      used as a baseline of bar charts for \texttt{k1c2} (blue) and \texttt{CRNN} (green). In other words, blue and green bar heights represent the performance gaps, \texttt{k2c1}-\texttt{k1c2} and \texttt{CRNN}-\texttt{k2c1}, respectively. \textit{iii}) Tags are grouped by categories (genre/mood/instrument/era) and sorted by the score of \texttt{k2c1}. \textit{iv}) The number in 
    parentheses after each tag indicates that tag's popularity ranking in the dataset.}
    \label{fig:auc_per_class}
  \vspace{-0.3cm}
\end{figure*}
% ↑  ---- FIGURE - AUC per tags ---- ↑

% ===== SUBSECTION: MEMORY ===== %
\vspace{-0.4cm}
\subsection{Memory-controlled experiment} \label{sec:mem}
\vspace{-0.2cm}
Figure \ref{fig:results} shows the AUCs for each network against the number of parameters. With the same number of parameters, the ranking of AUC is \texttt{CRNN} $>$ \texttt{k2c2} $>$ \texttt{k1c2} $>$\texttt{k2c1}. This indicates that \texttt{CRNN} can be preferred when the bottleneck is memory usage.
% For 3M-parameter structures, \texttt{Conv2D} outperforms \texttt{CRNN} while both of them outperform the state-of-the-art structure \cite{choi2016automatic}.

% ↓ ---- FIGURE - AUC by time---- ↓
\begin{figure}[t]
  \begin{center}
        \includegraphics[width=1.0\columnwidth]{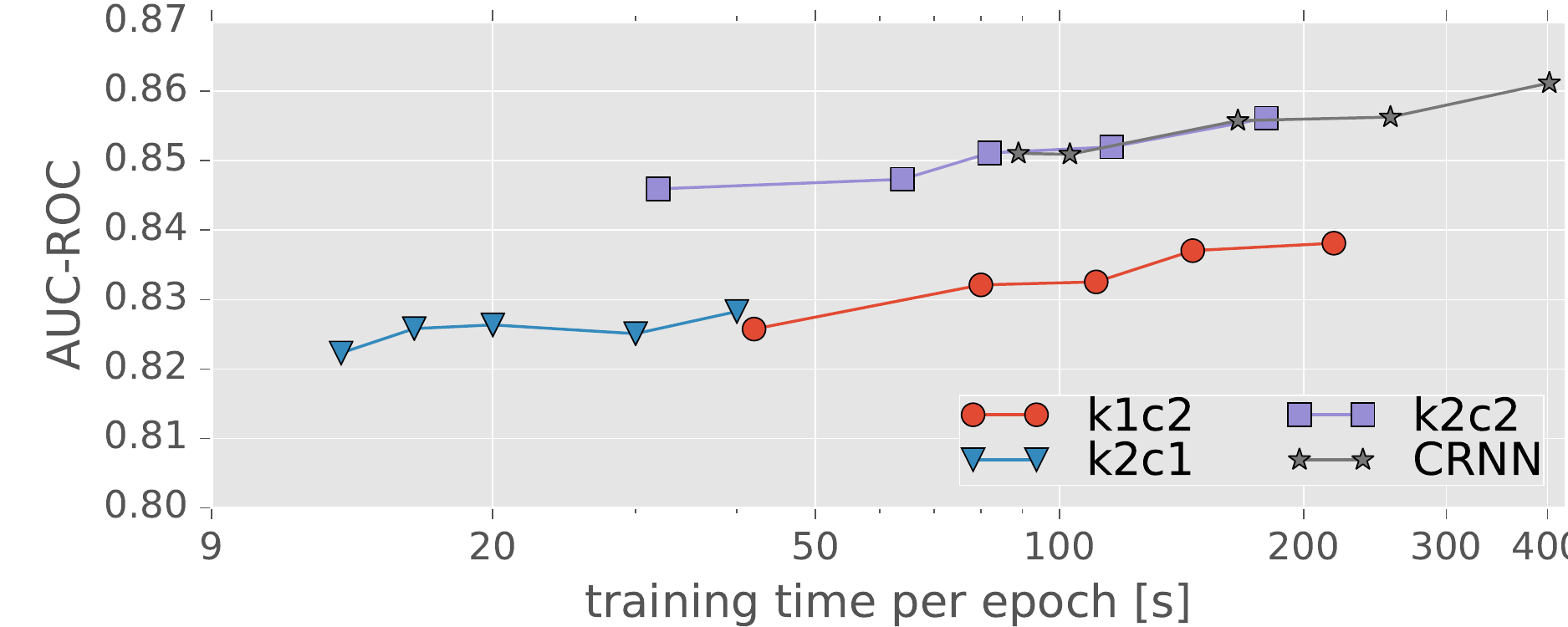}
             \vspace{-0.4cm}
        \caption{AUCs of the structures in training time - AUC plane. Each plot represents four different parameters, $\{$0.1, 0.25, 0.5, 1.0, 3.0$\}\times10^6$, from left to right.}\label{fig:results_time}
  \end{center}
  \vspace{-0.5cm}
\end{figure}
% ↑  ---- FIGURE - AUC by time ---- ↑

\texttt{CRNN} outperforms \texttt{k2c2} in all cases. Because they share the same 2D-convolutional layers, this difference is probably a consequence of the difference in RNNs and CNNs the ability of summarising the features over time. This may indicate that learning a global structure is more important than focusing on local structures for summarisation. One may focus on the different layer widths of two structures -- because recurrent layers use less parameters than convolutional layers, \texttt{CRNN} has wider convolutional layers than \texttt{k2x2} with same number of parameters. However, even \texttt{CRNN} with narrower layer widths (0.1M parameters) shows better performance than \texttt{k2c2} with wider widths (0.25M parameters). 

\texttt{k2c2} shows higher AUCs than \texttt{k2c1} and \texttt{k1c2} in all cases. This shows that the model of \texttt{k2c2}, which encodes local invariance and captures local time-frequency relationships, is more effective than the others, which ignores local frequency relationships. \texttt{k2c2} also uses parameters in a more flexible way with its fully-convolutional structure, while \texttt{k2c1} and \texttt{k1c2} allocate only a small proportion of the parameters to the feature extraction stage. For example, in \texttt{k1c2} with 0.5M parameters, only 13$\%$ of the parameters are used by convolutional layers while the rest, 87$\%$, are used by the fully-connected layers.

\texttt{k2c2} structures ($>$0.5M parameters) shows better performances than a similar but vastly larger structure in \cite{choi2016automatic}, which is shown as state of the art in Figure \ref{fig:results}. This is because the reduction in the number of feature maps removes redundancy.

The flexibility of \texttt{k1c2} may contribute the performance improvement over \texttt{k2c1}. In \texttt{k2c1}, the \textit{tall} 2-dimensional kernels in the first layer of \texttt{k2c1} compress the information of the whole frequency-axis pattern into each feature map. The following kernels then deal with this compressed representation with temporal convolutional and pooling. On the other hands, in \texttt{k1c2}, 1-dimensional kernels are shared over time and frequency axis until the end of convolutional layers. In other words, it gradually compress the information in time axis first, while preserving the frequency-axis pattern.

% ===== SUBSECTION: COMPUTATION ===== %
\vspace{-0.3cm}
\subsection{Computation-controlled comparison}
\vspace{-0.1cm}
We further investigate the computational complexity of each structure. The computational complexity is directly related to the training and prediction time and varies depending not only on the number of parameters but also on the structure. The wall-clock training times for 2500 samples are summarised in Table \ref{table:numbers} and plotted in Figure \ref{fig:results}.

The input compression in \texttt{k2c1} results in a fast computation, making it merely overlaps in time with other structures. The time consumptions of the other structures range in a overlapping region.

Overall, with similar training time, \texttt{k2c2} and \texttt{CRNN} show the best performance. 
% For example, the training speed of 0.1M-parameter \texttt{CRNN} is quicker than that of 0.5M-parameter \texttt{Conv2d} or 0.5M and 1M-parameter \texttt{Conv2d} and showing better performance. 
%\texttt{k2c2} networks are significantly faster However,  \texttt{Conv2D} with training time of 180s (3M parameters) outperforms \texttt{CRNN} with similar or longer training times (0.5M, 1M, and 3M parameters). 
This result indicates that either \texttt{k2c2} or \texttt{CRNN} can be used depending on the target time budget. 

With the same number of parameters, the ranking of training speed is always \texttt{k2c1} $>$ \texttt{k2c2} $>$ \texttt{k1c2} $>$ \texttt{CRNN}. There seems two factors that affect this ranking. \textbf{First}, among CNN structures, the sizes of feature maps are the most critical since the number of convolution operations is in proportion to the sizes. \texttt{k2c1} reduces the size of feature map in the first convolutional layer, where the whole frequency bins are compressed into one. \texttt{k2c2} reduces the sizes of feature maps in both axes and is faster than \texttt{k1c2} which reduces the sizes only in temporal axis. \textbf{Second}, the difference between \texttt{CRNN} and CNN structures arises from the negative correlation of speed and the depth of networks. The depth of \texttt{CRNN} structure is up to 20 (15 time steps in RNN and 5 convolutional layers), introducing heavier computation than the other CNN structures.

% ===== SUBSECTION: PER CLASS ===== %
\vspace{-0.3cm}
\subsection{Performance per tag}
\vspace{-0.1cm}
Figure \ref{fig:auc_per_class} visualises the AUC score of each tag of 1M-parameter structures. Each tag is categorised as one of genres, moods, instruments and eras, and sorted by AUC within its category. Under this categorisation, music tagging task can be considered as a multiple-task problem equivalent to four classification tasks with these four categories.

The \texttt{CRNN} outperforms \texttt{k2c1} for 44 tags, and \texttt{k2c1} outperforms \texttt{k1c2} for 48 out of 50 tags. From the multiple-task classification perspective, this result indicates that a structure that outperforms in one of the four tasks may perform best in the other tasks as well.

Although the dataset is imbalanced, the tag popularity (number of occurrence of each tag) is not correlated to the performance. Spearman rank correlation between tag popularity and the ranking of AUC scores of all tags is 0.077. It means that the networks effectively learn features that can be shared to predict different tags.
%--------------------------%
% Conclusions
%--------------------------%
\vspace{-0.3cm}
\section{Conclusions} \label{sec:con}
\vspace{-0.1cm}
We proposed a convolutional recurrent neural network (CRNN) for music tagging. In the experiment, we controlled the size of the networks by varying the numbers of parameters to for memory-controlled and computation-controlled comparison. Our experiments revealed that 2D convolution with 2d kernels (\texttt{k2c2}) and \texttt{CRNN} perform comparably to each other with a modest number of parameters. With a very small or large number of parameters, we observed a trade-off between speed and memory. The computation of \texttt{k2c2} is faster than that of \texttt{CRNN} across all parameter settings, while the \texttt{CRNN} tends to outperform it with the same number of parameters. 
% Future work will investigate RNN-based structures and audio input requirements for deep learning approaches. 

\bibliographystyle{IEEEbib}
\bibliography{icassp2017_auto_tagging_analysis}

\end{document}